\documentclass[11pt]{article}
\usepackage{amssymb, amsthm, amsmath}
\usepackage{graphicx}
\usepackage{bm}
\usepackage[authoryear]{natbib}
\usepackage{times}
\usepackage{float}
\usepackage{mathtools}
\usepackage[justification=centering]{caption}
\usepackage{enumerate}
\usepackage[inline]{enumitem}
\usepackage{setspace}
\usepackage{color}
\usepackage{xcolor}
\usepackage{amsmath,graphicx,centernot}
\usepackage{titling}
\providecommand{\keywords}[1]{\textbf{\textit{Keywords---}} #1}

\let\code=\texttt

\mathchardef\breakingcomma\mathcode`\,
{\catcode`,=\active
	\gdef,{\breakingcomma\discretionary{}{}{}}
}
\newcommand{\mathlist}[1]{$\mathcode`\,=\string"8000 #1$}

\newcommand{\bbeta}{\bm{\beta}}
\newcommand{\btheta}{\bm{\theta}}

\newcommand{\bX}{\bm{X}}

\newcommand{\bp}{\bm{p}}

\newcommand{\bz}{\bm{z}}

\newcommand{\bsig}{\bm{\sigma}}

\newtheorem{theorem}{Theorem}

\newlist{inlinelist}{enumerate*}{1}
\setlist*[inlinelist,1]{%
  label=(\roman*),
}

\begin{document}

\begin{center}
  {\Large {\bf Deep Distribution Regression}}\\ \vspace{12pt}
{\large {\bf Rui Li\textsuperscript{1}, Howard D. Bondell\textsuperscript{1,2} and Brian J. Reich\textsuperscript{1}}}\\ \vspace{6pt}
{\large {\textsuperscript{1}Department of Statistics, North Carolina State University}}\\ \vspace{4pt}
{\large {\textsuperscript{2}School of Mathematics and Statistics, University of Melbourne}}\\ \vspace{4pt}
\end{center}

\begin{abstract}
Due to their flexibility and predictive performance, machine-learning based regression methods have become an important tool for predictive modeling and forecasting. However, most methods focus on estimating the conditional mean or specific quantiles of the target quantity and do not provide the full conditional distribution, which contains uncertainty information that might be crucial for decision making. In this article, we provide a general solution by transforming a conditional distribution estimation problem into a constrained multi-class classification problem, in which tools such as deep neural networks. We propose a novel joint binary cross-entropy loss function to accomplish this goal. We demonstrate its performance in various simulation studies comparing to state-of-the-art competing methods. Additionally, our method shows improved accuracy in a probabilistic solar energy forecasting problem. 
\end{abstract}

\section{Introduction}\label{s:intro}
In recent years, a variety of machine learning methods, such as random forest, gradient boosting trees and neural networks have gained popularity and been widely adopted. These methods are often flexible enough to uncover complex relationships in high-dimensional data without strong assumptions on the underlying data structure. Off-the-shelf software is available to put these algorithms into production [\cite{pedregosa2011scikit}, \cite{45381} and \cite{paszke2017automatic}]. However, in regression and forecasting tasks, many of the machine learning methods only provide a point estimate,  without any additional information regarding the uncertainty of the target quantity. Understanding uncertainties are often crucial in fields such as financial markets and risk analysis [\cite{diebold1997evaluating}, \cite{timmermann2000density}], population and demographic studies [\cite{wilson2007probabilistic}], transportation and traffic analysis [\cite{zhu2017deep}, \cite{rodrigues2018beyond}] and energy forecasting [\cite{hong2016probabilistic}]. In this article, we establish a framework that can directly extend off-the-shelf machine learning algorithms to provide the full conditional distribution of the response given the covariates. Instead of estimating specific quantiles [\cite{koenker2001quantile}, \cite{taylor2000quantile} and \cite{friedman2001greedy}], or prediction intervals [\cite{khosravi2011comprehensive}, \cite{shrestha2006machine}], we aim to directly estimate the full conditional distribution, as other quantities can be directly extracted from it. 

This research is motivated by the necessity of probabilistic prediction in energy forecasting.   The Global Energy Forecasting Competition 2014 [\cite{hong2016probabilistic}] focused on probabilistic forecasting, due to the high demand of uncertainty estimation in energy forecasting, yet few existing methods are readily available. The solar energy forecasting track in this competition aimed to estimate the full conditional distribution of solar energy generation based on covariates such as solar radiance, temperature, time of the day, etc.  This is a crucial task in actual day-to-day operation as solar energy generation is highly weather dependent and thus volatile. To ensure stability in the electrical grid, grid operation not only requires accurate point prediction of solar energy generation, but also its volatility based on weather conditions. Driven by this practical problem, we establish a method to provide comprehensive information regarding the energy generation uncertainties, by estimating the full conditional distribution. Our method shows superior estimation accuracy in our real data analysis compared to competing methods. 

There are a number of approaches to estimate the conditional distribution of the target quantity $Y$ given the input vector $\bX$. A major class of methods estimates the density functions $f(Y, \bX)$ and $f(\bX)$ through kernel density estimators [\cite{Rosenblatt1969}] to obtain the conditional distribution estimate $f(Y|\bX) = f(Y, \bX)/f(\bX)$. This approach is limited by the dimensionality of the input space $\bX$, due to having to estimate the full joint distribution of $\bX$. Several methods have then been proposed to address some of the limitations in both locally parametric and non-parametric ways [\cite{hyndman1996estimating}, \cite{hyndman2002nonparametric}, \cite{holmes2012fast}, \cite{fan2009approximating}, \cite{izbicki2016nonparametric}]. Another popular approach is to approximate the distribution of interest by a parametric distribution family or mixture of such distributions, such as a mixture of Gaussians [\cite{escobar1995bayesian}, \cite{song2004density}, \cite{rojas2005conditional}, \cite{fahey2007conditional}]. This approach faces the challenge of determining the number of mixtures, and the approximation performance will depend on the complexity of the true underlying distribution. Bootstrap-based aggregation provides an alternative method, and a good example is quantile regression forest [\cite{meinshausen2006quantile}], which obtains the full conditional distribution from the empirical distribution in the aggregated tree leaves. A boosting approach to this problem was discussed in \cite{schapire2002modeling}. 

Machine learning approaches have experienced major success in classification problems. We leverage this success to build conditional distribution estimates. In our paper, we propose a two-stage method for conditional density estimation. In the first stage, we partition the response space into bins; in the second stage, the probabilities that the target variable $Y$ falls into each bin given the input covariates $\bX$ are estimated. In this way, we transform a conditional density estimation problem into a multi-class classification task, where we can take advantage of many off-the-shelf machine learning algorithms. This framework enjoys the model-agnostic property in the second stage, allows for plugging in any suitable multi-class classification method, such as deep learning for example. To further accommodate the fact that the classes are ordered, we use a joint binary cross entropy loss to couple with the neural network model. The design of our neural network also ensures the monotonicity of the estimated cumulative distribution function, which is not guaranteed by other ordinal classification methods as in \cite{frank2001simple} and \cite{cheng2008neural}. In addition, we explore random partitioning in the first stage followed by the ensemble approach to obtain a smoother and more stable density estimator. 

The paper is organized as follows: In Section \ref{s:hist}, we describe the model set up. In Section \ref{s:class}, we examine approaches and loss functions that can be used in the multi-class classification stage and in Section \ref{randhis} we examine an alternative partition method and model ensemble. In Section \ref{DEC}, we show that given the consistency of the classification model, we can achieve consistency for the density estimator.  We evaluate the model performance with simulation studies in Section \ref{Simu}. In the simulation study, we thoroughly examine the effect of number of bins, different binning strategies as well as different loss functions. In Section \ref{realD}, the method is illustrated using the aforementioned solar energy example where we demonstrate superior performance to quantile random forests. We conclude with the discussion in Section \ref{Discussion}.

\section{Distribution Estimation by Partitioning}\label{s:hist}
Our approach is based on partitioning the range of the response variable $Y$ into bins and approximating the conditional density function $f(y|\bX)$ by a piecewise constant function. We also propose ensemble random partitioning which allows for a smooth density. Formally, assume we are interested in the density of $Y$ in the range of $[l, u]$. This interval is partitioned by $m$ cut-points $l < c_1 < c_2 < \cdots < c_m <u$ into $m+1$ bins. Let $c_0=l, c_{m+1}=u$ and $T_i = [c_{i-1}, c_{i})$ denotes the $i$th bin, for $i \in \lbrace 1, \cdots, m+1 \rbrace$. 

Given the independent variables $\bX$ and the partition, denote the $p_i(\bX) = P(Y \in T_i|\bX)$ for $i=1, 2, \cdots, m+1$ as the conditional probability that $Y$ belongs to the $i$th interval of the partition. Assume the density function in the $i$th bin can be approximated by the constant function $f(y|\bX) = p_i(\bX)/|T_i|$, for all $y \in T_i$, then the conditional density function has the form:

\begin{gather}\label{dens-est}
f(y|\bX) \approx \sum_{i=1}^{m+1}\frac{p_i(\bX)}{|T_i|}I(y\in T_i), 
\end{gather}
where $|T_i| = c_i - c_{i-1}$ is the size of the interval $T_i$. 

We then estimate $p_i(\bX)$ with a classification model. Let $\hat{p}_i(\bX)$ be the estimator for $p_i(\bX)$ obtained from the classification model. Plugging $\hat{p}_i(\bX)$ into (\ref{dens-est}) gives the density estimator: 

\begin{gather}
\hat{f}(y|\bX) = \sum_{i=1}^{m+1}\frac{\hat{p}_i(\bX)}{|T_i|}I(y\in T_i).
\end{gather}
A natural approach to estimate $p_i(\bX)$ is discussed in Section \ref{s:class}. 

\subsection{Probability Estimation for Each Partitioned Bin}\label{s:class}
Below we describe two different modeling strategies for the multi-class classification task of estimating $p_i(\bX)$.  
\subsubsection{Multinomial Log-likelihood}\label{s:MLL}
A natural approach to obtain the estimates for the conditional probability of a response observation in each bin is to model the conditional distribution as a multinomial distribution. We note that this does not take into account the fact that the bins are ordered. We will discuss how to deal with this fact in the next section. For a given observation $\bX_n$, where $n = 1, 2, \cdots, N$, we let $\bm{G}_n = \bm{e}_i$ represent the event $Y_n \in T_i$, where $\bm{e}_i$ is the vector of zeros except for $1$ in the $i$th element. We then assume that $\bm{G}_n|\bX_n \sim Multinomial(1, \bp(\bX_n;\btheta))$, where $ $\mathlist{\bp(\bX_n;\btheta) = [p_1(\bX_n;\btheta), p_2(\bX_n;\btheta), \cdots, p_{m+1}(\bX_n;\btheta)]}$ $ is the probability vector describing the conditional probability of $Y_n$ belonging to each bin given $\bX_n$, and $\btheta$ denotes the parameters in the classification model. The log-likelihood function is:

\begin{gather}
\sum_{n=1}^N\log\mathcal{L}(\btheta|Y_n, \bX_n) = \sum_{n=1}^N\sum_{i=1}^{m+1} I(Y_n \in T_i)\log[p_i(\bX_n, \btheta)].
\end{gather}

Given an appropriate classification model $\bp(\bX;\btheta)$, our goal is to maximize the log-likelihood function with respect to $\btheta$. In this work, we use deep neural networks as a flexible and robust method. Under this model, $\btheta$ corresponds to its biases and weights. The softmax function $\bsig[\bz(\bX_n;\btheta)]_i = \bm{e}^{z_i}/\sum_{j=1}^{m+1}\bm{e}^{z_j}$ was used as the activation function for the output layer of the neural network model, where $\bz(\bX_n;\btheta)$ is the output layer vector prior to the softmax transformation. The softmax activation function ensures that the estimator $\hat{\bp}(\bX_n;\btheta)$ is a valid probability vector, such that $\hat{p}_i(\bX_n;\btheta) = \bsig[\bz(\bX_n;\btheta)]_i > 0$ and $\sum_{i=1}^{m+1}\hat{p}_i(\bX_n;\btheta)=1$. 

Although we choose the neural network model, our framework is model-agnostic, and any method that is appropriate for multi-class classification can be utilized. This approach is very straight-forward in practice as many off-the-shelf machine learning tools have multi-class classification algorithms implemented. However, one potential caveat is that this approach can be sensitive to the number of cut-points $m$. While larger $m$ is required to uncover the fine details of the target distribution and reduce bias, it results in increased variance due to smaller number of observations per partitioned bin. Additionally, this approach does not take the order of the bins into consideration. In order to address some of these concerns, we provide a modified method in Section \ref{s:EBCL}. 

\subsubsection{Joint Binary Cross Entropy Loss}\label{s:EBCL}
Instead of utilizing multi-class classification to directly estimate the function $\bp(\bX)$, we reframe the problem as $m$ binary classification tasks, and binary classifications are evaluated at each of the $m$ cut-points. In other words, we will obtain the conditional cumulative distribution function $F(c_j;\bX_n, \btheta) = P(Y_n \leq c_j|\bX_n)$ by estimating $F(c_1;\bX_n, \btheta),F(c_2;\bX_n, \btheta), \cdots,F(c_m;\bX_n, \btheta)$ jointly.

In this approach, we retain the model (i.e. neural networks) for the conditional probability that $Y_n$ belongs to each bin given $\bX_n$, but specify the loss function in terms of the cumulative distribution function $F(c_j;\bX_n, \btheta) = \sum_{i=1}^{j}p_i(\bX_n; \btheta)$. Because the estimates $\hat{p}_i(\bX_n; \btheta)$ are positive, we are able to automatically obtain the monotonicity property of $F(c_j;\bX_n, \btheta)$. For a given cut-point $c_j$, the binary cross entropy (BCE) loss is:

\begin{align}
BCE(c_j) &= -\sum_{n=1}^N \lbrace I(Y_n \leq c_j)\log [F(c_j;\bX_n, \btheta)] \nonumber
\\&+ [1 - I(Y_n \leq c_j)]\log [1-F(c_j;\bX_n, \btheta)]\rbrace.
\end{align}
Combining the BCEs across all $m$ cut-points gives the joint binary cross entropy (JBCE) loss:

\begin{align}
JBCE &= \sum_{j=1}^mBCE(c_j).
\end{align}

Our goal becomes minimizing the JBCE loss with respect to $\btheta$. This approach takes the order of the bins into consideration in contrast to multi-class classification where the relationships among the bins are ignored.  In scenarios where the number of cut-points $m$ is large, the number of observations per bin can be small and the estimation of $p_i(\bX_n;\btheta)$ can be poor in the multi-class classification setting. Thus this alternative approach can provide an advantage in that it will remain stable even for larger number of cut-points. This will be seen in our simulation study.

The concept is motivated by the ordinal classification approach proposed by \cite{frank2001simple}, yet we provide two additional advantages over the original method. First, instead of building $m$ independent binary classification models, we model the $m$ binary classification events jointly, which greatly reduces the computational cost. Secondly, our model ensures monotonicity of the estimated cumulative distribution function while the original method by \cite{frank2001simple} does not have this guarantee. 

\subsection{Distribution Estimation with Ensemble Random Partitioning}\label{randhis}
Approximating the conditional density with a piecewise constant function can suffer from two drawbacks: 
\begin {enumerate*} [label=(\arabic*)]
\item The results are sensitive to the choice of the cut-points, and 
\item The density or the cumulative distribution functions are not smooth.
\end {enumerate*} 
In order to address the two issues, we propose an ensemble random partitioning method. This method fits $K$ independent density estimators, each of which has different positions of the $m$ cut-points and the final ensemble estimator is an average over the $K$ individual estimators. For the $k$th estimator, we generate $v_{ik} \overset{\text{iid}}{\sim} \mbox{Uniform}(l, u)$ for $i \in \lbrace 1, \cdots, m \rbrace$ and $k \in \lbrace 1, \cdots, K \rbrace$. Let $c_{ik}$ be the $i$th smallest value in $\lbrace v_{1k}, v_{2k}, \cdots, v_{mk}\rbrace$ so that $l<c_{1k}<c_{2k}<\cdots<c_{mk}<u$, and $c_{0k} = l, c_{m+1, k} = u$. Then the $i$th interval of the $k$th partition is $T_{ik} = [c_{i-1, k}, c_{ik})$. The random partition estimator is then defined as:

\begin{gather}
\hat{f}(y|\bX) = \frac{1}{K}\sum_{k=1}^{K}\sum_{i=1}^{m+1}\frac{\hat{p}_{ik}(\bX)}{|T_{ik}|}I(y\in T_{ik}),
\end{gather} 
where $\hat{p}_{ik}(\bX)$ is the estimator of $P(y \in T_{ik}|\bX)$ for the $k$th estimator.

The density estimator $\hat{f}(y|\bX)$ is an average from $K$ random partitioning estimators. This estimator alleviates the necessity of choosing the cut-points locations and can approximate a smooth function as $K \to \infty$. The trade-off is that the computation time increases linearly in $K$.

\section{Density Estimation Consistency} \label{DEC}

In this section, we study the conditions that are required for consistent density estimation. We denote the density estimator as $\hat{f}(y|\bX)$, and consider the estimator to be consistent if the integrated mean squared error $\int_{u}^{l}[\hat{f}(y|\bX) - f(y|\bX)]^2dy \rightarrow 0$ asymptotically.

\begin{theorem}
	Let $f(y|\bX)$ be the target conditional density, and ${\cal P}_1, \cdots, {\cal P}_K$ be equally sized, consecutive and non-overlapping bins on the true density support $[l,u]$. The true probability in bin $k$ is $\pi_k(\bX) = \int_{{\cal P}_k}f(y|\bX)dy$. If the target density and classification estimator $\hat{\pi}_k(\bX)$ follow conditions (i) to (iv)  , then the density estimator $\hat{f}_K(y|\bX) = \sum_{k=1}^K\frac{1}{|{\cal P}_k|}I(y\in{\cal P}_k){\hat \pi}_k(\bX)$ is consistent.
	\begin{enumerate}[label=(\roman*)]
		\item $\forall y, \bX$, $f(y|\bX) < \infty, |f'(y|\bX)| < \infty$ and $|f''(y|\bX)| < \infty$.
		\item As $n \rightarrow \infty$, $K \rightarrow \infty$ and $K/n \rightarrow 0$.
		\item $Bias(\hat{\pi}_k(\bX)) = o(1/K), \forall k \in {1, 2, \cdots, K}$.
		\item $Var(\hat{\pi}_k(\bX)) = o(1/K^2), \forall k \in {1, 2, \cdots, K}$.
	\end{enumerate}
\end{theorem}

The proof of Theorem 1 is given in the appendix. This theorem is agnostic to the classification model. Note that we are not aware of results showing that deep learning estimation of $\hat{\pi}_k(\bX)$ obtains properties (iii) and (iv). However, for illustration purposes, here we show as an example that multi-class logistic regression satisfies conditions (iii) and (iv), given a proper choice of class number $K$. 

Let $h = {\cal P}_k$ denote the bin width and assume we are interested in the target density on a finite support, such that $-\infty < l < u < +\infty$. The input vector $\bX$ has the dimensionality $p$. For the logistic regression model, $\pi_k(\bX) = \sigma(\bX^T\bbeta_k) = \bm{e}^{\bX^T\bbeta_k}/(1 + \sum_{j=1}^{K-1}\bm{e}^{\bX^T\bbeta_j})$, where the coefficient $\bbeta_k$ has dimensionality $p$ and let $\bbeta$ denote the full parameter $[\bbeta_1^T, \cdots, \bbeta_{K-1}^T]^T$ with dimensionality $(K-1)p$. Thus the bias for $\pi_k(\bX)$ is:

\begin{align}
E[\hat{\sigma}(\bX^T\hat{\bbeta}_k) - \sigma(\bX^T\bbeta_k)] &=  E\bigg[\frac{d\sigma(\bX^T\bbeta_k)}{d\bbeta}\Bigr|_{\bbeta=\bbeta^\ast}(\hat{\bbeta} - \bbeta)\bigg] \nonumber \\
&= -\sum_{j=1 , j \ne k}^{K-1}E[\sigma(\bX^T\bbeta^\ast_j)\sigma(\bX^T\bbeta^\ast_k)\bX^T(\hat{\bbeta}_j - \bbeta_j)] \nonumber \\
&+ E[\sigma(\bX^T\bbeta^\ast_k)\lbrace 1 - \sigma(\bX^T\bbeta^\ast_k) \rbrace \bX^T (\hat{\bbeta}_k - \bbeta_k)].
\end{align}
By the mean value theorem, $\bbeta^\ast$ is a point between $\hat{\bbeta}$ and the true value $\bbeta$. Based on results from \cite{he2000parameters}, given $K\log (K)/n \rightarrow 0 $, we have $\lVert\hat{\bbeta}-\bbeta\rVert^2 = O_P(K/n)$, thus $\sup_{k}\lVert\hat{\bbeta}_k-\bbeta_k\rVert^2 = O_P(K/n)$, and $\sup_{k}E\lVert\hat{\bbeta}_k - \bbeta_k\rVert_2 = O(\sqrt{K/n})$. By condition $(i)$, the density is finite everywhere, we have $\forall j, \sigma(\bX^T\bbeta_j) = O(1/K)$ and $\sigma(\bX^T\bbeta^\ast_j) = O_P(1/K)$. Thus from $(7)$, we can obtain the $\sup_{k}E[\hat{\sigma}(\bX^T\hat{\bbeta}_k) - \sigma(\bX^T\bbeta_k)] = O(\sqrt{1/(Kn)})$. 

The Taylor expansion of $\hat{\sigma}(\bX^T\hat{\bbeta}_k)$ is :

\begin{align}
\hat{\sigma}(\bX^T\hat{\bbeta}_k) = \sigma(\bX^T\bbeta_k) + \frac{d\sigma(\bX^T\bbeta_k)}{d\bbeta}(\hat{\bbeta} - \bbeta) + o(K/n).
\end{align}
\cite{he2000parameters} showed that $\alpha^T(\hat{\bbeta} - \bbeta)$ is asymptotically normal with $Var(\alpha^T(\hat{\bbeta} - \bbeta)) = O(\lVert \alpha \rVert^2/n)$, if $K^2(\log K)/n \rightarrow 0$, where in this case $\alpha = \frac{d\sigma(\bX^T\bbeta_k)}{d\bbeta}$. Similar to the derivation in (7), we can obtain $\lVert \frac{d\sigma(\bX^T\bbeta_k)}{d\bbeta} \rVert^2 = O(1/K^2)$, and thus $Var(\hat{\sigma}(\bX^T\hat{\bbeta}_k)) =  O(1/nK^2)$. 

Given $K^2(\log K)/n \rightarrow 0$, we have that both bias and variance meet conditions (iii) and (iv).

\section{Simulation Study}\label{Simu}
We conduct a simulation study to examine the distribution estimation accuracy of our method comparing to quantile regression forest (QRF), originally proposed by \cite{meinshausen2006quantile}. Quantile regression forest is increasingly being used in the energy and weather forecasting field [\cite{taillardat2016calibrated}, \cite{van2018review}], due to its robustness, flexibility and mature implementation in R (\code{quantregForest}) and Python (\code{scikit-garden}). 

We applied our method to data generated from a variety of distributions. The scenarios we have tested are a linear model with normally distributed errors (Model 1), mixture distributions with nonlinear mean functions (Models 2 and 3) and a skewed distribution with nonlinear mean function (Model 4). The model specification details are:

\begin{enumerate}
	\item Model 1:  $Y = \bX^T\bbeta_1 + \exp(\bX^T\bbeta_2)*\epsilon$, where $X_1, \cdots, X_5 \overset{iid}{\sim} N(0, 1), \epsilon \sim N(0, 1), \bbeta_1 \sim N(\bm{0},I_5), \bbeta_2 \sim N(\bm{0},0.45I_5). $
	\item Model 2:  $Y = [10sin(2\pi X_1 X_2) + 10X_4 + \epsilon_1]\pi_1 + [20(X_3-0.5)^2 + 5X_5 + \epsilon_2](1-\pi_1)$, where $X_1,\cdots, X_{10} \overset{iid}{\sim} \mbox{Uniform}(0,1), \pi_1 \sim \mbox{Bernoulli}(0.5), \epsilon_1 \sim N(0, 2.25), \epsilon_2 \sim N(0,1).$
	\item Model 3:  $[sin(X_1) + \epsilon_1]\pi_1 + [2sin(1.5X_1+1) + \epsilon_2](1-\pi_1)$, where $\quad X_1 \sim Uniform(0,10), \pi_1 \sim Bernoulli(0.5), \epsilon_1 \sim N(0, 0.09),\epsilon_2 \sim N(0,0.64).$
	\item Model 4:  $Y = 10sin(2\pi X_1 X_2) + 20(X_3-0.5)^2 + 10X_4  + 5X_5 + \epsilon $, where $X_1,\cdots, X_{10} \overset{iid}{\sim} Uniform(0,1), \epsilon \sim SkewNormal(0,1,-5).$
\end{enumerate}

Model performances are evaluated using several proper scoring rules summarized by \cite{gneiting2007strictly}. Specifically, the scoring rules that are used here are:

\begin{enumerate}
	\item Continuous ranked probability score (CRPS): 
	\begin{align}\label{CRPS}
	CRPS = \frac{1}{N}\sum\limits_{n=1}^{N}\int_{l}^{u}\lbrace \hat{F}(y|\bX_n)-I(y\geq Y_n)\rbrace^2dy.
	\end{align}
	The integral is approximated by summing over $1000$ evenly spaced grid points between $l$ and $u$, and normalized by its range. Lower CRPS score indicates better model performance. 
	\item Average quantile loss (AQTL):
	\begin{gather}\label{AQTL}
	QTL(\tau) = \frac{1}{N}\sum\limits_{n=1}^{N} [(Y_n - \hat{Q}(\tau|\bX_n))(\tau - I\lbrace Y_n \leq \hat{Q}(\tau|\bX_n) \rbrace)] \\
	AQTL = \frac{1}{99}\sum_{t=1}^{99} QTL(t/100).
	\end{gather}
	$\hat{Q}(\tau|\bX_n)$ is the estimated quantile function (derived from $\hat{f}(y|\bX_n)$) and AQTL is the averaged quantile loss over $99$ percentiles. AQTL is also the evaluation metric used in GEFCom2014 [\cite{hong2016probabilistic}]. Lower AQTL indicates better model performance. 
	\item Empirical coverage of the $90\%$ prediction interval $[\hat{Q}(0.05|\bX_n), \hat{Q}(0.95|\bX_n)]$.
\end{enumerate}

In all simulations, the neural network model are constructed using \code{TensorFlow} and \code{Keras}. It has three hidden layers with 100 neurons each. The exponential linear unit (ELU) activation function is used at each hidden layer and the softmax activation function is used at the output layer. Regularization via dropout is also applied [\cite{srivastava2014dropout}], with dropout rate at $50\%$ for each hidden layer. The quantile regression forest model is built with \code{scikit-garden}, with $500$ trees per forest. In addition, we also include the logistic regression as an alternative multi-class classification method to compare with the neural network model. For both the neural network and logistic regression, we evaluated their performance across different number of partitioned bins. We also examine the performance with different objective functions for the neural network model, as described in Section \ref{s:class}. The effect of ensemble random partitioning is also evaluated. The ensemble model consisted of $K=20$ independent density estimators as described in Section \ref{randhis}.  

\begin{figure}[!htb]
	\begin{minipage}{\textwidth}
		\centering
		\includegraphics[width=0.76\textwidth]{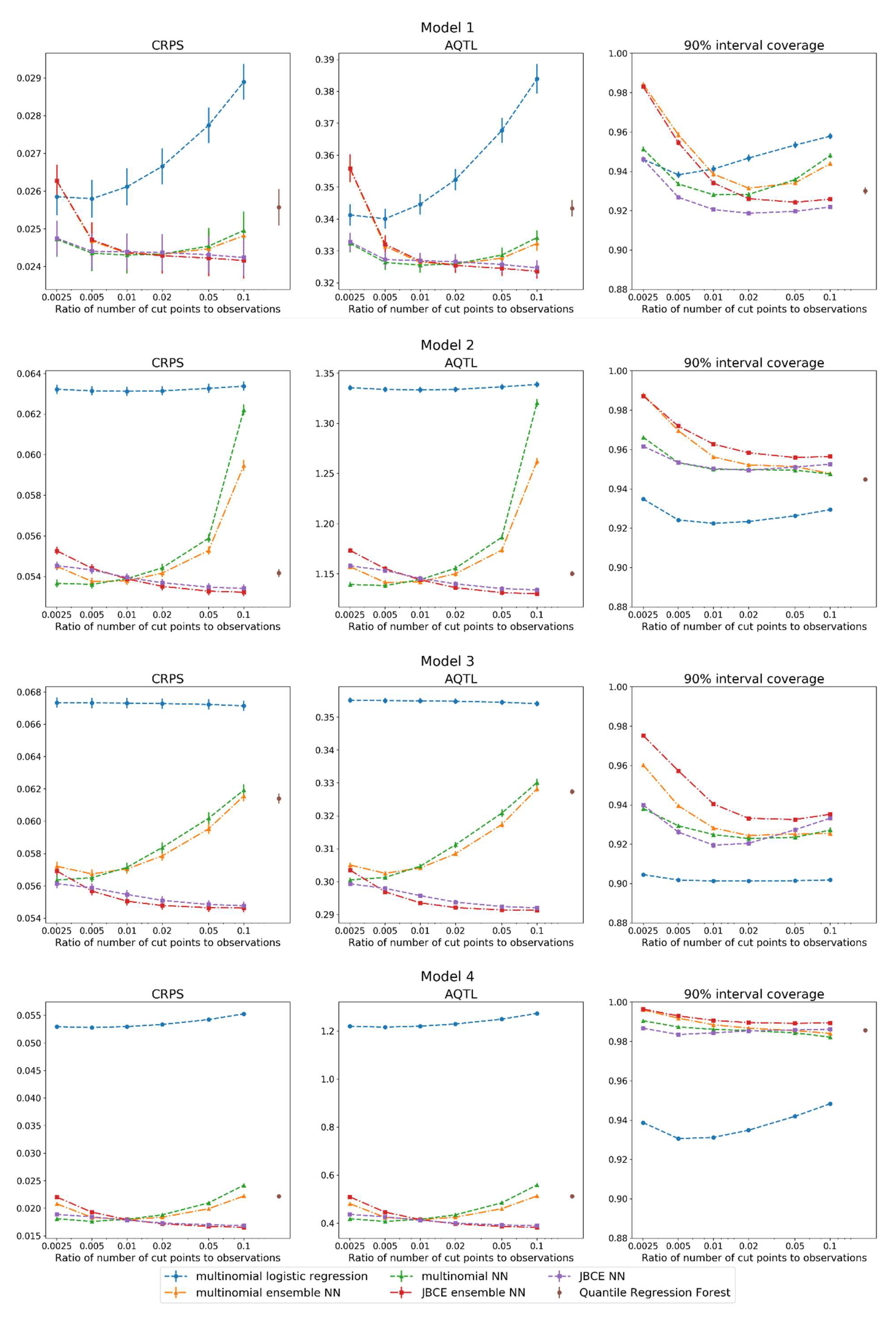}
	\end{minipage}
	\caption{Model Comparison in Simulations}\label{m1sim}
   \medskip
\parbox{\linewidth}{\small%
	\textbf{multinomial logistic regression}: Logistic regression as the multi-class classification model, with evenly spaced fixed partition. \textbf{multinomial ensemble NN}: Ensemble model consisted of $20$ neural network models trained with multinomial log-likelihood as the objective function. \textbf{multinomial NN}: Neural network model trained with multinomial log-likelihood as objective function, with evenly spaced fixed partition. \textbf{JBCE ensemble NN}: Ensemble model consisted of $20$ neural network models trained with JBCE loss as objective function. \textbf{JBCE NN}: Neural network model trained with JBCE loss as objective function, with evenly spaced fixed partition. \textbf{CRPS}: Continuous ranked probability score. \textbf{AQTL}: Average quantile loss.} 
\end{figure}

The simulations are run over $100$ independent datasets, each with $6000$ observations in the training set and $1000$ observations in the testing set, on which evaluations are compared. The results are summarized in Figure \ref{m1sim}. 

In all the tested scenarios, using the neural network shows superior performance over logistic regression. Additionally, the neural network with JBCE loss demonstrated superior performance over the quantile regression forest. With the JBCE loss, increasing the number of partitions results in stable improvement of the performance, while with the multinomial log-likelihood as objective function, the performance is sensitive to the number of partitions and requires fine-tuning. Ensemble random partitioning provides slight improvements for the neural network approach. Note that the random partitioning will produce smoothed estimates which would not be achieved with fixed partitioning.

\section{Application to the GEFCOM2014 Dataset}\label{realD}
The Global Energy Forecasting Competition 2014 was an IEEE sponsored competition that focused on probabilistic forecasting [\cite{hong2016probabilistic}] and attracted hundreds of teams and individuals. It had four competition tracks: electricity load forecasting, electricity price forecasting, solar energy forecasting and wind energy forecasting. We applied our method to the solar energy forecasting dataset and compared its performance with quantile regression forest, which was widely used among the top teams. 

The dataset has two years of data from three solar farms in Australia (no other spatial information is available). The solar power outputs were collected at hourly frequency, together with 12 weather forecast variables from the European Centre for Medium-range Weather Forecasts (ECMWF), which include surface solar radiance,  net solar radiation, surface thermal radiation, temperature, total cloud coverage, precipitation, total column liquid water, total column ice water, relative humidity, surface pressure, 10-meter U wind component and 10-meter V wind component. All weather forecast variables are provided as point forecasts. The solar power output was scaled by the competition organizers to the range of [0,1], so $l=0$ and $u=1$. The forecasting task is to use these weather forecast variables to predict the conditional distribution of solar power output for holdout months. 

\begin{figure}[H]
	\begin{minipage}{\textwidth}
		\centering
		\includegraphics[width=.95\textwidth]{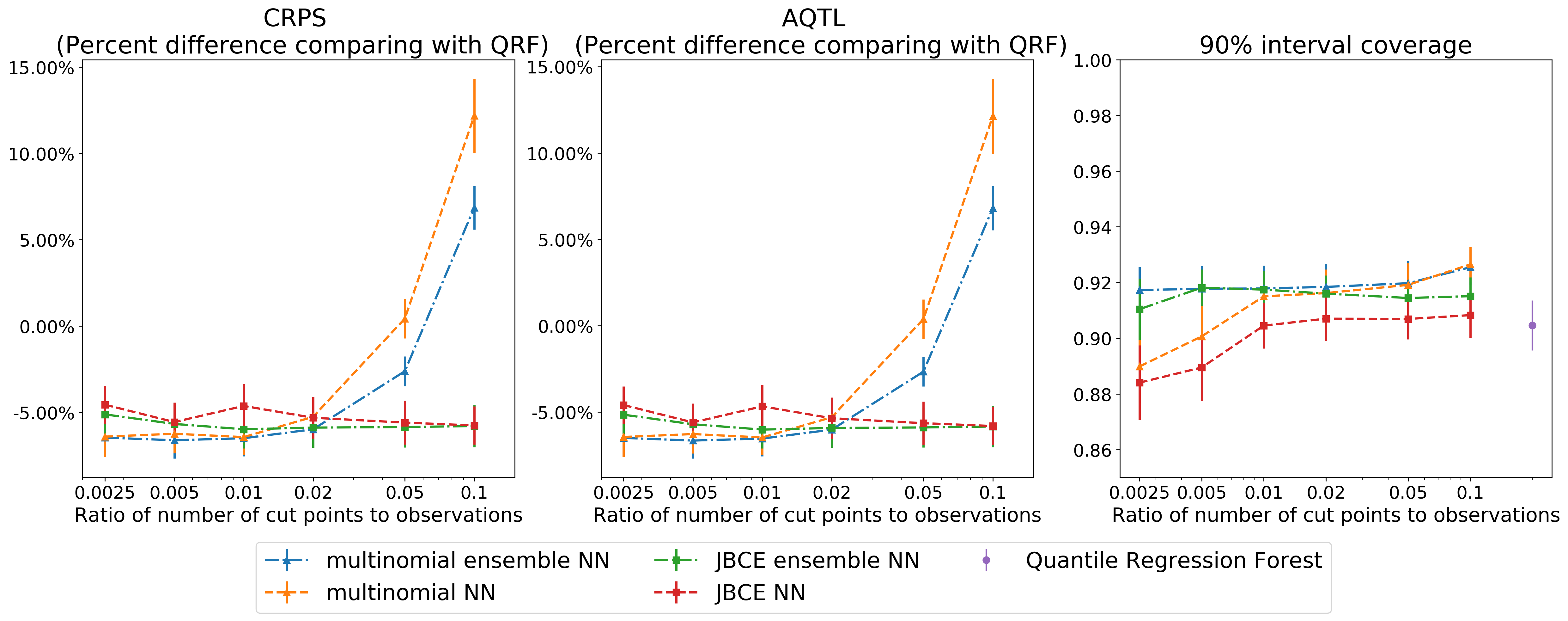}
	\end{minipage}
	\caption{Model Comparison with the Solar Energy Generation Datasets}\label{realdata}
   \medskip
\parbox{\linewidth}{\small%
	\textbf{multinomial ensemble NN}: Ensemble model consisted of $20$ neural network models trained with multinomial log-likelihood as the objective function. \textbf{multinomial NN}: Neural network model trained with multinomial log-likelihood as objective function, with evenly spaced fixed partition. \textbf{JBCE ensemble NN}: Ensemble model consisted of $20$ neural network models trained with JBCE loss as objective function. \textbf{JBCE NN}: Neural network model trained with JBCE loss as objective function, with evenly spaced fixed partition. \textbf{CRPS}: Continuous ranked probability score. \textbf{AQTL}: Average quantile loss. \textbf{QRF}: Quantile regression forest. 
} 
\end{figure}

We preprocessed the data to include additional variables that are commonly used in solar energy forecasting, such as indicator variables for each solar farm and hour of the day. We also created sine and cosine transformations for the date of the year, as solar energy output shows strong seasonality. We conducted the experiments to compare our method with quantile regression forest in a rolling simulation setting: We used the first year of data as the initial training dataset, and used the immediate next month as the testing set for evaluation. Then we appended the testing dataset to the initial training set to retrain the models, and the model performances were re-evaluated on the next month as the new testing set. This process was repeated for 12 hold out months on which the models were evaluated. The results are included in Figure \ref{realdata}. We compared different model specifications with the quantile regression forest. The y-axis of the first two panels are the percent change in CRPS or AQTL by comparing with the quantile regression forest. The percent change is evaluated for each testing month and averaged across months. Lower scores indicate better performance. 

Our method with JBCE loss consistently achieves about a $5\%$ reduction in CRPS or AQTL compared to the quantile regression forest. However, with the multinomial log-likelihood objective function, the performance is sensitive to the number of cut points, and in this dataset, a smaller number of cut points is preferred when using multinomial log-likelihood objective function. This result is consistent with our simulations that models trained with JBCE loss are less sensitive to the number of cut points, and thus require less tuning. 

\begin{figure}[H]
	\begin{minipage}{\textwidth}
		\centering
		\includegraphics[width=.9\textwidth]{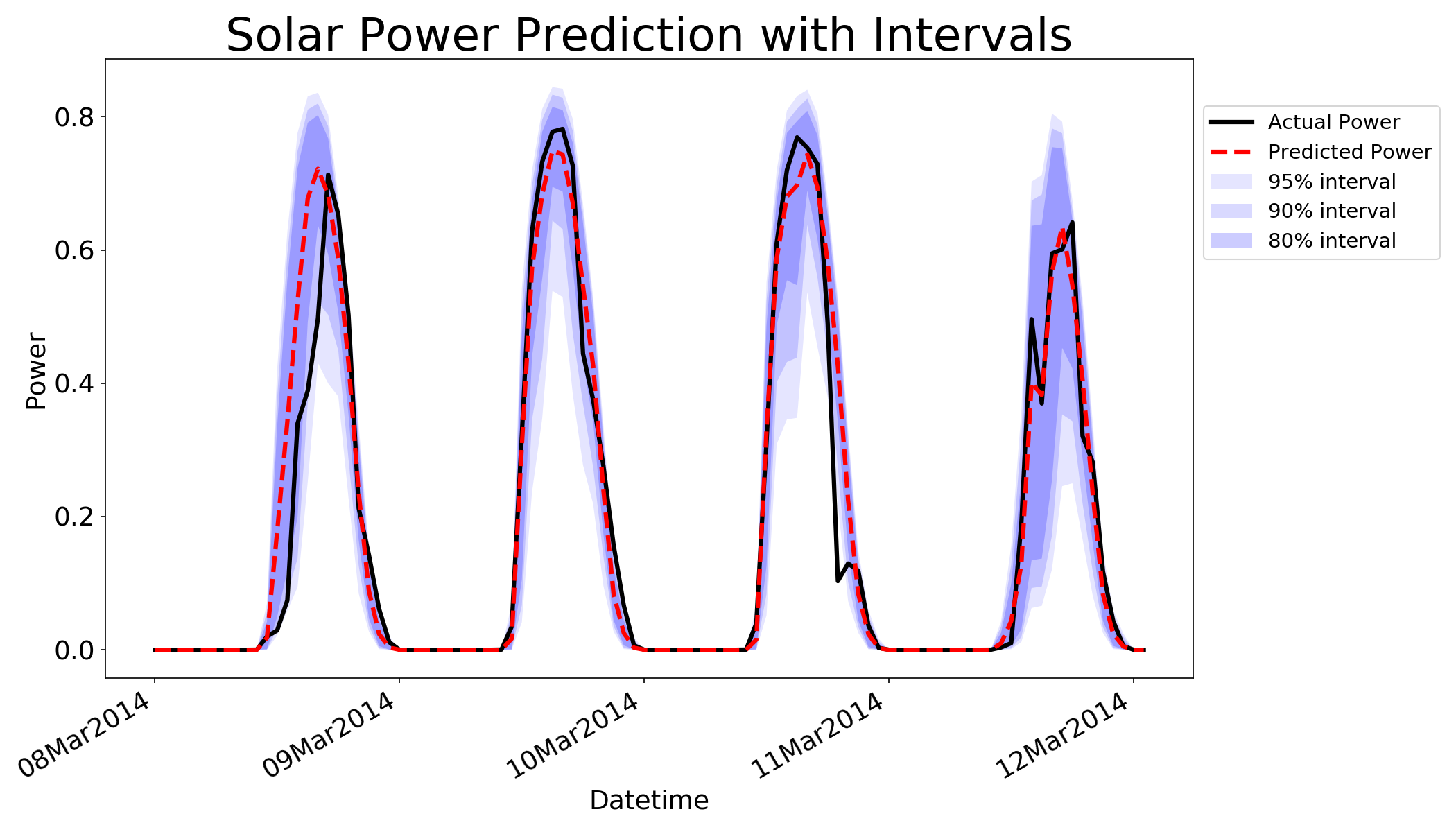}
	\end{minipage}
	\caption{Solar Power Prediction with Intervals}\label{solar_farm}
   \medskip
\parbox{\linewidth}{\small%
	The solar power prediction for solar farm 1 in the datasets are plotted for the first five days of March, 2014. The prediction method is using the ensemble model consisted of $20$ neural network models trained with the JBCE loss. } 
\end{figure}

In Figure \ref{solar_farm}, we show an example of the distribution pattern of solar energy output revealed by our model. We only focus on the time period during the day, since the solar energy at night is always zero and does not require forecasting. For days such as March 10th, 2014, the weather is clear, and the prediction intervals are narrow, and thus the predicted energy output has little variability. The predicted distribution on days forecasted to be sunny is often negatively skewed, as energy output cannot be higher than its maximum capacity, yet can be negatively impacted by unexpected cloudy conditions. On March 12th, 2014, wide prediction intervals indicate volatility in solar energy output. This information helps grid operators prepare for unexpected fluctuations in energy output, where point forecasts fail to do so. 

\begin{figure}[H]
	\begin{minipage}{\textwidth}
		\centering
		\includegraphics[width=.47\textwidth]{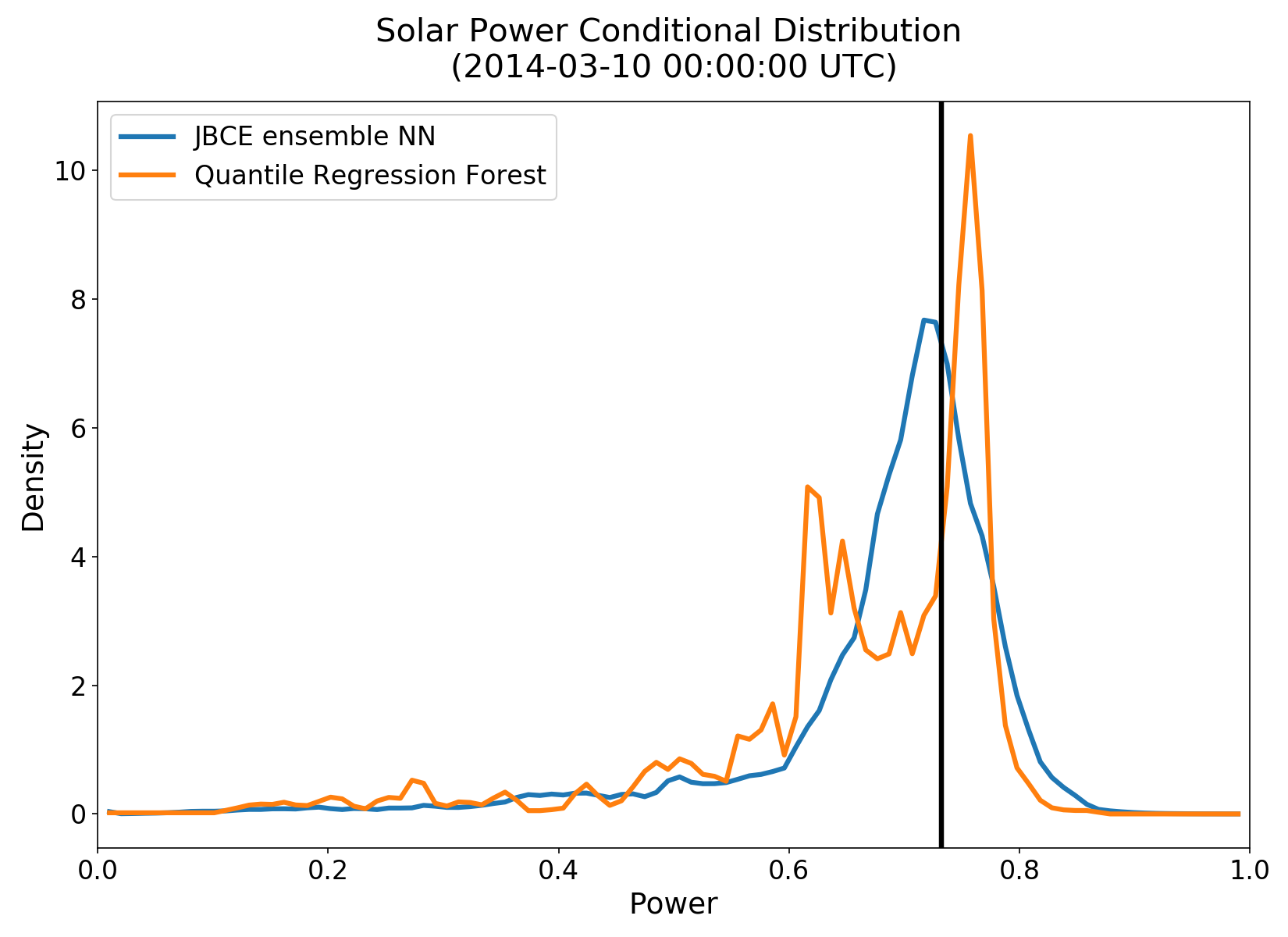}
		\includegraphics[width=.47\textwidth]{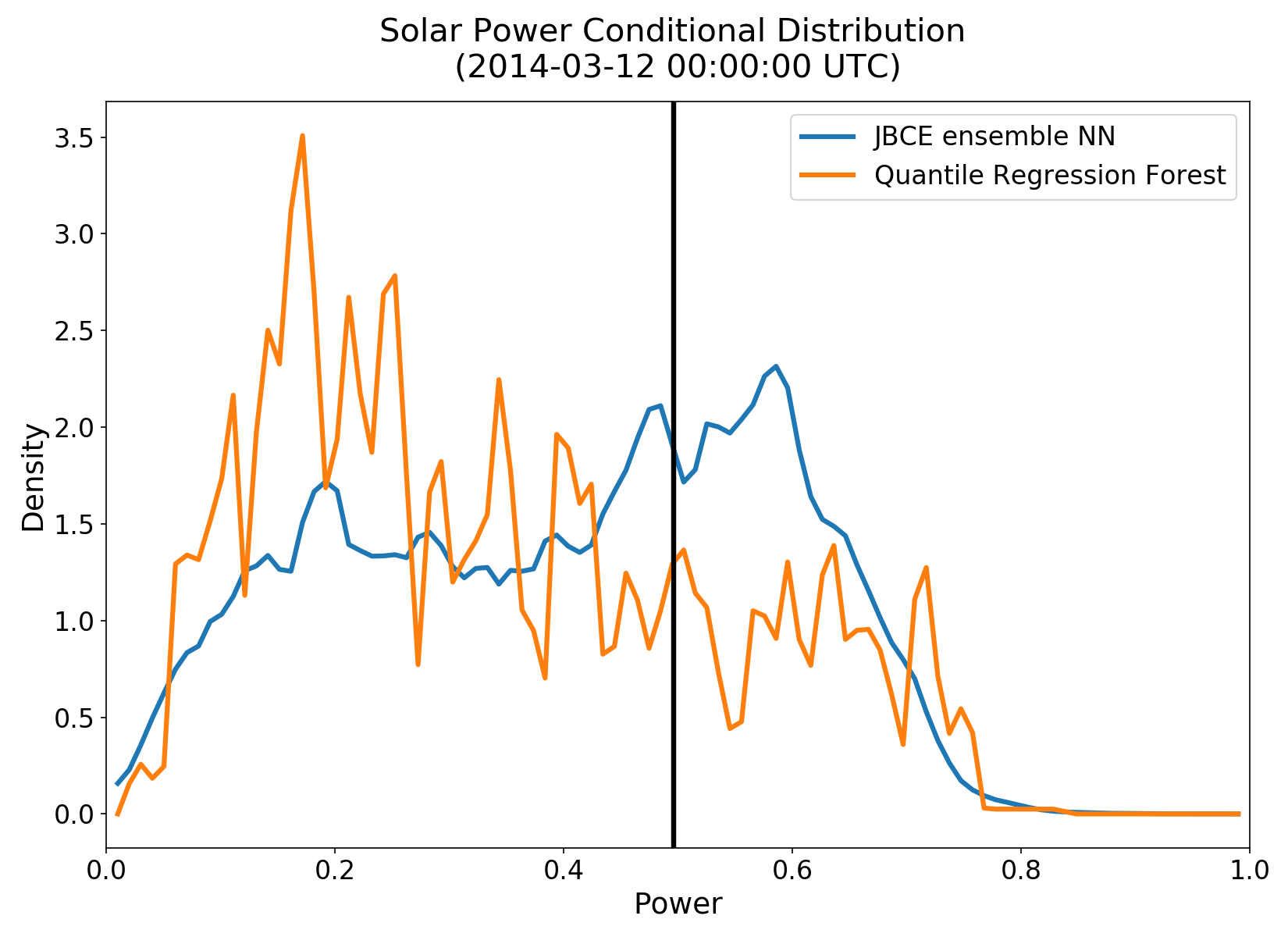}
	\end{minipage}
	\caption{Solar Power Conditional Density}\label{solar_density}
	  \medskip
	\parbox{\linewidth}{\small%
		\textbf{JBCE ensemble NN}: Ensemble model consisted of $20$ neural network models trained with JBCE loss as objective function. The densities are evaluated at $100$ evenly spaced grid points in $[0,1]$. For a given grid point, the density is derived from the estimated cumulative distribution function $\hat{f}(y|\bX) = \lbrace \hat{F}(y+h/2|\bX)-\hat{F}(y-h/2|\bX)\rbrace /h$, where $h$ is the grid width.} 
\end{figure}

We zoom into March 10th and March 12th, 0:00 AM UTC of Figure \ref{solar_farm} and show the estimated conditional density in Figure \ref{solar_density}. March 10th represents a sunny day and shows a relatively concentrated density with slight negative skewness, whereas March 12th represents a day with fluctuating weather and a wide conditional density. The densities are evaluated at $100$ evenly spaced grid points in $[0,1]$. Though both methods generate non-smooth cumulative distribution functions, and thus spiky density functions, our method with ensemble strategy shows a much smoother pattern.

\section{Discussion} \label{Discussion}
In this paper, we leverage the success of machine learning classification models such as neural network to build conditional distribution estimates. Our proposed two-stage framework transforms the conditional distribution estimation problem into a multi-class classification problem, which allows us to use flexible and robust machine learning tools. To thoroughly understand this framework, we examined the effect of different partitioning methods, classification models, and objective functions. We also established theoretical foundations for our framework to achieve consistent estimation. Our results revealed that a neural network trained with joint binary cross entropy loss can achieve superior performance without the sensitivity to the number of partitions. In both simulations and the solar energy generation dataset, we showed our model can obtain better performance than popular quantile regression forest. With the ability to provide full conditional distribution, our model can give useful insight in solar energy forecasting practices.

This research is focused on harnessing the power of machine learning methods for uncertainty estimation. We believe our research opens up opportunities for bringing in machine learning models into distribution estimation problems. 

\section{Acknowledgements}
The authors' work was partially supported by King Abdullah University of Science and Technology (grant number 3800.2).

\newpage
\section{Appendix}
Following is the proof for the Theorem 1. This proof is related to the proof of histogram consistency as in Theorem 6.11 of \cite{wasserman2006all}, with the consideration of the bias and variance from the classification model. Let $p_k$ denote $\pi_k(\bX)$, $b_k$ denote $Bias(\hat{\pi}_k(\bX))$ and $v_k$ denote $Var(\hat{\pi}_k(\bX))$. Without the loss of generality, let's assume the range of $y$ is $[0, 1]$. Since $K$ denotes the number of equally sized, consecutive and non-overlapping bins, let $h$ denote the bin width, and we have $Kh=1$.  For a given $y$ in the support of $f(y|\bX)$, $\exists k \in 1, 2, \cdots, K$, such that $y \in {\cal P}_k$. Then we have

\begin{align}
p_k =& \int_{{\cal P}_k}f(u|\bX)du \nonumber \\
= &\int_{{\cal P}_k}\lbrace f(y|\bX) + (u-y)f'(y|\bX) + \frac{(u-y)^2}{2}f''(\tilde{y}|\bX)\rbrace du \nonumber \\
= &hf(y|\bX) + hf'(y|\bX)(y_0 - y) + \mathcal{O}(h^3) \nonumber 
\end{align}

Where $\tilde{y}$ is between $u$ and $y$. $y_0$ is the middle point of interval ${\cal P}_k$. 

So the bias $b(y|\bX)$ for the density estimator is following:
\begin{align}
b(y|\bX) =& E[\frac{\hat{p_k}}{h}] - f(y|\bX) \nonumber \\
= &\frac{p_k + b_k}{h}-f(y|\bX) \nonumber \\
= &\frac{b_k}{h} + f'(y|\bX)(y_0 - y) + \mathcal{O}(h^2) \nonumber 
\end{align}
Integrate $b^2(y|\bX)$ over the interval ${\cal P}_k$:
\begin{align}
\int_{{\cal P}_k}b^2(y|\bX) \leq & 2\int_{{\cal P}_k} \frac{b_k^2}{h^2} + \lbrace f'(y|\bX)(y_0 - y)\rbrace^2dy  + \mathcal{O}(h^5)\nonumber \\
= & 2\frac{b_k^2}{h} + 2f'(\tilde{y_k}|\bX)^2\int_{{\cal P}_k}(y_0-y)^2dy + \mathcal{O}(h^5) \nonumber \\
= & 2\frac{b_k^2}{h} + f'(\tilde{y_k}|\bX)^2\frac{h^3}{6} + \mathcal{O}(h^5) \nonumber 
\end{align}
Where $\tilde{y_k} \in {\cal P}_k$. 

We then integrate $b^2(y|\bX)$ over the range of $y$.
\begin{align}
\int_{0}^{1}b^2(y|\bX)dy =& \sum_{k=1}^{K}\int_{{\cal P}_k}b^2(y|\bX)dy\nonumber \\
\leq &\sum_{k=1}^{K}\lbrace 2\frac{b_k^2}{h} + f'(\tilde{y_k}|\bX)^2\frac{h^3}{6} \rbrace + \mathcal{O}(h^4)\nonumber \\
= &2\sum_{k=1}^{K}\frac{b_k^2}{h} + \frac{h^2}{6}\sum_{k=1}^{K}hf'(\tilde{y_k}|\bX)^2 + \mathcal{O}(h^4) \nonumber \\
= &2\sum_{k=1}^{K}\frac{b_k^2}{h} + \frac{h^2}{6}\int_{0}^{1}f'(y|\bX)^2dy + \mathcal{O}(h^4) \nonumber
\end{align}

If the model bias $b_k = o(h) = o(1/K), \forall k \in 1, 2, \cdots, K$, we can achieve $\int_{0}^{1}b^2(y|\bX)dy \to 0$ as $K \to \infty$.

For the variance part: 
\begin{align}
\int_{0}^{1}Var(\hat{f}(y|\bX)) =& \sum_{k=1}^{K}\int_{{\cal P}_k}Var(\frac{\hat{p_k}}{h}) =\sum_{k=1}^{K}\int_{{\cal P}_k}\frac{v_k}{h^2}\nonumber \\
=& \sum_{k=1}^{K}\frac{v_k}{h}\nonumber 
\end{align}

If the model variance $v_k = o(h^2) = o(1/K^2), \forall k \in 1, 2, \cdots, K$, we can achieve $\int_{0}^{1}Var(\hat{f}(y|\bX))dy \to 0$ as $K \to \infty$. 

\newpage
\begin{singlespace}
	\bibliographystyle{rss}
	\bibliography{Template}

\begin{thebibliography}{34}
\expandafter\ifx\csname natexlab\endcsname\relax\def\natexlab#1{#1}\fi
\expandafter\ifx\csname url\endcsname\relax
  \def\url#1{\texttt{#1}}\fi
\expandafter\ifx\csname urlprefix\endcsname\relax\def\urlprefix{URL}\fi

\bibitem[{Abadi et~al.(2016)Abadi, Barham, Chen, Chen, Davis, Dean, Devin,
  Ghemawat, Irving, Isard, Kudlur, Levenberg, Monga, Moore, Murray, Steiner,
  Tucker, Vasudevan, Warden, Wicke, Yu and Zheng}]{45381}
Abadi, M., Barham, P., Chen, J., Chen, Z., Davis, A., Dean, J., Devin, M.,
  Ghemawat, S., Irving, G., Isard, M., Kudlur, M., Levenberg, J., Monga, R.,
  Moore, S., Murray, D.~G., Steiner, B., Tucker, P., Vasudevan, V., Warden, P.,
  Wicke, M., Yu, Y. and Zheng, X. (2016) Tensorflow: A system for large-scale
  machine learning.
\newblock In \textit{12th USENIX Symposium on Operating Systems Design and
  Implementation (OSDI 16)}, 265--283.
\newblock
  \urlprefix\url{https://www.usenix.org/system/files/conference/osdi16/osdi16-abadi.pdf}.

\bibitem[{Cheng et~al.(2008)Cheng, Wang and Pollastri}]{cheng2008neural}
Cheng, J., Wang, Z. and Pollastri, G. (2008) A neural network approach to
  ordinal regression.
\newblock In \textit{2008 IEEE International Joint Conference on Neural
  Networks (IEEE World Congress on Computational Intelligence)}, 1279--1284.
  IEEE.

\bibitem[{Diebold et~al.(1997)Diebold, Gunther and Tay}]{diebold1997evaluating}
Diebold, F.~X., Gunther, T.~A. and Tay, A. (1997) Evaluating density forecasts.

\bibitem[{Escobar and West(1995)}]{escobar1995bayesian}
Escobar, M.~D. and West, M. (1995) Bayesian density estimation and inference
  using mixtures.
\newblock \textit{Journal of the American Statistical Association},
  \textbf{90}, 577--588.

\bibitem[{Fahey et~al.(2007)Fahey, Thane, Bramwell and
  Coward}]{fahey2007conditional}
Fahey, M.~T., Thane, C.~W., Bramwell, G.~D. and Coward, W.~A. (2007)
  Conditional gaussian mixture modelling for dietary pattern analysis.
\newblock \textit{Journal of the Royal Statistical Society: Series A
  (Statistics in Society)}, \textbf{170}, 149--166.

\bibitem[{Fan et~al.(2009)Fan, Peng, Yao and Zhang}]{fan2009approximating}
Fan, J.-q., Peng, L., Yao, Q.-w. and Zhang, W.-y. (2009) Approximating
  conditional density functions using dimension reduction.
\newblock \textit{Acta Mathematicae Applicatae Sinica, English Series},
  \textbf{25}, 445--456.

\bibitem[{Frank and Hall(2001)}]{frank2001simple}
Frank, E. and Hall, M. (2001) A simple approach to ordinal classification.
\newblock In \textit{European Conference on Machine Learning}, 145--156.
  Springer.

\bibitem[{Friedman(2001)}]{friedman2001greedy}
Friedman, J.~H. (2001) Greedy function approximation: a gradient boosting
  machine.
\newblock \textit{Annals of statistics}, 1189--1232.

\bibitem[{Gneiting and Raftery(2007)}]{gneiting2007strictly}
Gneiting, T. and Raftery, A.~E. (2007) Strictly proper scoring rules,
  prediction, and estimation.
\newblock \textit{Journal of the American Statistical Association},
  \textbf{102}, 359--378.

\bibitem[{He and Shao(2000)}]{he2000parameters}
He, X. and Shao, Q.-M. (2000) On parameters of increasing dimensions.
\newblock \textit{Journal of Multivariate Analysis}, \textbf{73}, 120--135.

\bibitem[{Holmes et~al.(2012)Holmes, Gray and Isbell}]{holmes2012fast}
Holmes, M.~P., Gray, A.~G. and Isbell, C.~L. (2012) Fast nonparametric
  conditional density estimation.
\newblock \textit{arXiv preprint arXiv:1206.5278}.

\bibitem[{Hong et~al.(2016)Hong, Pinson, Fan, Zareipour, Troccoli and
  Hyndman}]{hong2016probabilistic}
Hong, T., Pinson, P., Fan, S., Zareipour, H., Troccoli, A. and Hyndman, R.~J.
  (2016) Probabilistic energy forecasting: Global energy forecasting
  competition 2014 and beyond.
\newblock \textit{International Journal of Forecasting}, \textbf{32}, 896--913.

\bibitem[{Hyndman et~al.(1996)Hyndman, Bashtannyk and
  Grunwald}]{hyndman1996estimating}
Hyndman, R.~J., Bashtannyk, D.~M. and Grunwald, G.~K. (1996) Estimating and
  visualizing conditional densities.
\newblock \textit{Journal of Computational and Graphical Statistics},
  \textbf{5}, 315--336.

\bibitem[{Hyndman and Yao(2002)}]{hyndman2002nonparametric}
Hyndman, R.~J. and Yao, Q. (2002) Nonparametric estimation and symmetry tests
  for conditional density functions.
\newblock \textit{Journal of Nonparametric Statistics}, \textbf{14}, 259--278.

\bibitem[{Izbicki and Lee(2016)}]{izbicki2016nonparametric}
Izbicki, R. and Lee, A.~B. (2016) Nonparametric conditional density estimation
  in a high-dimensional regression setting.
\newblock \textit{Journal of Computational and Graphical Statistics},
  \textbf{25}, 1297--1316.

\bibitem[{Khosravi et~al.(2011)Khosravi, Nahavandi, Creighton and
  Atiya}]{khosravi2011comprehensive}
Khosravi, A., Nahavandi, S., Creighton, D. and Atiya, A.~F. (2011)
  Comprehensive review of neural network-based prediction intervals and new
  advances.
\newblock \textit{IEEE Transactions on Neural Networks}, \textbf{22},
  1341--1356.

\bibitem[{Koenker and Hallock(2001)}]{koenker2001quantile}
Koenker, R. and Hallock, K.~F. (2001) Quantile regression.
\newblock \textit{Journal of economic perspectives}, \textbf{15}, 143--156.

\bibitem[{Van~der Meer et~al.(2018)Van~der Meer, Wid{\'e}n and
  Munkhammar}]{van2018review}
Van~der Meer, D.~W., Wid{\'e}n, J. and Munkhammar, J. (2018) Review on
  probabilistic forecasting of photovoltaic power production and electricity
  consumption.
\newblock \textit{Renewable and Sustainable Energy Reviews}, \textbf{81},
  1484--1512.

\bibitem[{Meinshausen(2006)}]{meinshausen2006quantile}
Meinshausen, N. (2006) Quantile regression forests.
\newblock \textit{Journal of Machine Learning Research}, \textbf{7}, 983--999.

\bibitem[{Paszke et~al.(2017)Paszke, Gross, Chintala, Chanan, Yang, DeVito,
  Lin, Desmaison, Antiga and Lerer}]{paszke2017automatic}
Paszke, A., Gross, S., Chintala, S., Chanan, G., Yang, E., DeVito, Z., Lin, Z.,
  Desmaison, A., Antiga, L. and Lerer, A. (2017) Automatic differentiation in
  pytorch.
\newblock In \textit{NIPS-W}.

\bibitem[{Pedregosa et~al.(2011)Pedregosa, Varoquaux, Gramfort, Michel,
  Thirion, Grisel, Blondel, Prettenhofer, Weiss, Dubourg
  et~al.}]{pedregosa2011scikit}
Pedregosa, F., Varoquaux, G., Gramfort, A., Michel, V., Thirion, B., Grisel,
  O., Blondel, M., Prettenhofer, P., Weiss, R., Dubourg, V. et~al. (2011)
  Scikit-learn: Machine learning in python.
\newblock \textit{Journal of Machine Learning Research}, \textbf{12},
  2825--2830.

\bibitem[{Rodrigues and Pereira(2018)}]{rodrigues2018beyond}
Rodrigues, F. and Pereira, F.~C. (2018) Beyond expectation: Deep joint mean and
  quantile regression for spatio-temporal problems.
\newblock \textit{arXiv preprint arXiv:1808.08798}.

\bibitem[{Rojas et~al.(2005)Rojas, Genovese, Miller, Nichol and
  Wasserman}]{rojas2005conditional}
Rojas, A.~L., Genovese, C.~R., Miller, C.~J., Nichol, R. and Wasserman, L.
  (2005) Conditional density estimation using finite mixture models with an
  application to astrophysics.

\bibitem[{Rosenblatt(1969)}]{Rosenblatt1969}
Rosenblatt, M. (1969) Conditional probability density and regression
  estimators.
\newblock \textit{In P. R. Krishnaiah (Ed.), Multivariate analysis ii}, 25--31.

\bibitem[{Schapire et~al.(2002)Schapire, Stone, McAllester, Littman and
  Csirik}]{schapire2002modeling}
Schapire, R.~E., Stone, P., McAllester, D., Littman, M.~L. and Csirik, J.~A.
  (2002) Modeling auction price uncertainty using boosting-based conditional
  density estimation.
\newblock In \textit{ICML}, 546--553.

\bibitem[{Shrestha and Solomatine(2006)}]{shrestha2006machine}
Shrestha, D.~L. and Solomatine, D.~P. (2006) Machine learning approaches for
  estimation of prediction interval for the model output.
\newblock \textit{Neural Networks}, \textbf{19}, 225--235.

\bibitem[{Song et~al.(2004)Song, Yang and Pavel}]{song2004density}
Song, X., Yang, K. and Pavel, M. (2004) Density boosting for gaussian mixtures.
\newblock In \textit{International Conference on Neural Information
  Processing}, 508--515. Springer.

\bibitem[{Srivastava et~al.(2014)Srivastava, Hinton, Krizhevsky, Sutskever and
  Salakhutdinov}]{srivastava2014dropout}
Srivastava, N., Hinton, G., Krizhevsky, A., Sutskever, I. and Salakhutdinov, R.
  (2014) Dropout: a simple way to prevent neural networks from overfitting.
\newblock \textit{The Journal of Machine Learning Research}, \textbf{15},
  1929--1958.

\bibitem[{Taillardat et~al.(2016)Taillardat, Mestre, Zamo and
  Naveau}]{taillardat2016calibrated}
Taillardat, M., Mestre, O., Zamo, M. and Naveau, P. (2016) Calibrated ensemble
  forecasts using quantile regression forests and ensemble model output
  statistics.
\newblock \textit{Monthly Weather Review}, \textbf{144}, 2375--2393.

\bibitem[{Taylor(2000)}]{taylor2000quantile}
Taylor, J.~W. (2000) A quantile regression neural network approach to
  estimating the conditional density of multiperiod returns.
\newblock \textit{Journal of Forecasting}, \textbf{19}, 299--311.

\bibitem[{Timmermann(2000)}]{timmermann2000density}
Timmermann, A. (2000) Density forecasting in economics and finance.
\newblock \textit{Journal of Forecasting}, \textbf{19}, 231--234.

\bibitem[{Wasserman(2006)}]{wasserman2006all}
Wasserman, L. (2006) \textit{All of nonparametric statistics}.
\newblock Springer Science \& Business Media.

\bibitem[{Wilson and Bell(2007)}]{wilson2007probabilistic}
Wilson, T. and Bell, M. (2007) Probabilistic regional population forecasts: The
  example of queensland, australia.
\newblock \textit{Geographical Analysis}, \textbf{39}, 1--25.

\bibitem[{Zhu and Laptev(2017)}]{zhu2017deep}
Zhu, L. and Laptev, N. (2017) Deep and confident prediction for time series at
  uber.
\newblock In \textit{2017 IEEE International Conference on Data Mining
  Workshops (ICDMW)}, 103--110. IEEE.

\end{thebibliography}
\end{singlespace}

\end{document}